\documentclass{IOS-Book-Article}

\usepackage{mathptmx}
\usepackage{soul}
\setuldepth{article}

\usepackage{amssymb}
\usepackage{amsmath}
\usepackage{xurl}
\usepackage{graphicx}
\usepackage{algorithm}
\usepackage{algpseudocode}
\usepackage[most]{tcolorbox}
\usepackage{ragged2e}
\usepackage{booktabs}
\usepackage{hyperref}
\usepackage{caption}
\usepackage{subcaption}
\usepackage{makecell}

\makeatletter
\newcounter{alg@savedline}
\makeatother

\def\hb{\hbox to 11.5 cm{}}

\begin{document}
\pagestyle{headings}
\begin{frontmatter}              

\title{From Answers to Guidance: A Proactive Dialogue System for Legal Documents}


\author{\fnms{Ashish} \snm{Chouhan}
\thanks{Corresponding Author: Ashish Chouhan, chouhan@informatik.uni-heidelberg.de}} and
\author{\fnms{Michael} \snm{Gertz}
}

\address{Data Science Group, \\ Institute of Computer Science, \\Heidelberg University, Germany}

\begin{abstract}
The accessibility of legal information remains a constant challenge, particularly for laypersons seeking to understand and apply complex institutional texts. While the European Union provides open access to legislation, parliamentary responses, and regulatory documents, these resources can be challenging for laypeople to explore. In this paper, we introduce EUDial, a proactive multi-turn dialogue dataset constructed from 204 blogs curated by the Citizens’ Enquiries Unit (AskEP) of the European Parliamentary Research Service. EUDial contains 880 dialogue turns (averaging 4.3 turns per dialogue), where each dialogue includes initial questions, structured answers, and follow-up questions. Beyond dataset construction, we propose the LexGuide framework that leverages retrieval-augmented generation with hierarchical topic organization to structure dialogue progression, ensuring both comprehensive coverage of legal aspects and coherence across conversational turns. The results demonstrate that proactive, structured navigation closes the gap between the availability of legal information and citizen comprehension, establishing EUDial and LexGuide as practical resources for advancing proactive legal dialogue systems.
\end{abstract}

\begin{keyword}
legal natural language processing\sep dialogue dataset\sep European Union\sep follow-up questions\sep proactive dialogue system
\end{keyword}
\end{frontmatter}

\section{Introduction}
\label{sec:introduction}
Around 865k legal documents are accessible via the EUR-Lex platform\footnote{\url{https://eur-lex.europa.eu/statistics/eu-law-statistics.html}, \\ accessed on 1st September 2025} for the years 1990 to 2025, including legislation, case law, and treaties. The Citizens' Enquiries Unit (AskEP)\footnote{\url{https://epthinktank.eu/author/epanswers}, accessed on 2nd September 2025} within the European Parliamentary Research Service (EPRS) allows citizens to pose questions and receive responses. While AskEP responses are not legal documents, they serve as informative texts of public relevance. Additionally, the European Union (EU) Publications Office has introduced Publio\footnote{\url{https://op.europa.eu/en/web/webtools/publio-the-publications-office-virtual-assistant}, accessed on 2nd September 2025}, a virtual assistant for EU law access. Cherubini et al.~\cite{cherubini2024improving} describe the project Chat-EUR-Lex, a dialogue system for EUR-Lex regulations in English and Italian.

Despite these advances, a substantial gap remains between the availability of legal information and citizens' ability to comprehend and properly use that information. Legal texts are highly technical, characterized by complex terminology, intricate cross-references, and domain-specific reasoning patterns~\cite{ariai2024natural}. Consider a citizen seeking to understand healthcare rights in another EU member state. While Chat-EUR-Lex or Publio can respond, these are reactive dialogue systems~\cite{den-et-al-2025-survey}, where users lead dialogue and systems follow the user's instructions. Reactive systems heavily rely on the user's ability to formulate precise queries without being guided toward broader or related issues, such as exceptions or related directives.

Traditional dialogue\footnote{In this paper, the terms "dialogue" and "conversation" are used interchangeably.} systems share this limitation, operating primarily using a reactive paradigm~\cite{den-et-al-2025-survey}. While generating accurate responses to independent questions, they rarely anticipate the user's informational needs where knowledge is layered and interconnected. This problem is particularly present in legal contexts, where non-expert users (laypersons) may not know which follow-up questions to ask for a comprehensive understanding, potentially leaving critical aspects unexplored. Although dialogue systems have advanced in healthcare and education, the legal domain remains comparatively underexplored~\cite{mo2024survey}. Most legal natural language processing (NLP) research concentrates on document summarization, classification, or judgment prediction serving legal professionals~\cite{ariai2024natural, janatian2023text, aumiller-etal-2022-eur, chouhan-gertz-2024-lexdrafter}, while resources for laypersons have received limited attention~\cite{janatian2023text, buttner-habernal-2024-answering, askari2023closer}. Large language models (LLMs) demonstrate remarkable multi-turn interaction capabilities~\cite{mo2024survey}, but face challenges with complicated document structures, complex legal language, and heterogeneous legal documents~\cite{ariai2024natural, ganguly-et-al-2023-legalir, jiang-etal-2024-leveraging}.

In this work, we introduce EUDial (\textit{E}uropean \textit{U}nion \textit{Dial}ogues), a dialogue dataset generated by transforming AskEP blogs into multi-turn dialogues with queries, answers, and relevant follow-up questions. We also propose the LexGuide (Guided Exploration of Legal Documents) framework that leverages Retrieval-Augmented Generation (RAG)~\cite{lewis2020retrieval} to retrieve relevant information and construct hierarchical topic trees for structuring dialogue systems. Using educational scaffolding principles~\cite{vygotsky1978mind}, LexGuide helps transform dialogue systems from reactive dialogue systems to proactive dialogue systems that actively guide users through topic exploration. Our work makes two key contributions: First, we present a methodology that systematically transforms single-turn QA pairs from AskEP blogs into multi-turn dialogues, thereby creating the benchmark dataset EUDial for evaluating proactive dialogue systems. Second, we introduce the LexGuide framework, which formalizes dialogue progression as retrieval, hierarchical organization, and strategic navigation, thus enabling proactive behavior while maintaining topical coherence between dialogue turns\footnote{Code for EUDial dataset construction and LexGuide framework are available at \url{https://github.com/achouhan93/EUDial-LexGuide}}. The framework transitions from traditional reactive dialogue systems to intelligent conversation partners that can guide users through complex topics.

The paper is organized as follows: Section~\ref{sec:relatedwork} reviews prior research, Section~\ref{sec:datasetmodel} describes the EUDial dataset, and Section~\ref{sec:dialogue-framework} introduces the LexGuide framework. Sections~\ref{sec:experiment-setup} and~\ref{sec:evaluation} present the experimental setup and evaluation, and Section~\ref{sec:conclusion} concludes with key findings and future directions.

\section{Related Work}
\label{sec:relatedwork}
Our work sits at the intersection of dialogue systems, follow-up question generation, and legal NLP, specifically developing proactive dialogue systems that guide conversations and provide explanations to laypersons seeking legal information.

\paragraph{\textbf{Proactive Dialogue Systems.}} Traditional dialogue systems operate reactively, following user-driven conversations. However, this reactive paradigm fails when serving laypersons who lack domain expertise to formulate proper follow-up questions. Recent research addresses this limitation through proactive dialogue systems that actively guide users toward their information goals~\cite{den-et-al-2025-survey}. Zhang et al.~\cite{zhang-et-al-2025-medark} propose the ask and retrieve knowledge framework, where LLMs iteratively generate sub-queries for decision making. Butala et al.~\cite{butala-etal-2024-promise} propose ProMISe, where systems generate suggested questions targeting atomic aspects of a user's intent. While promising, these approaches have not been systematically applied to information-seeking contexts, where users require a structured understanding of complex topics, particularly in domains such as the legal domain.

Follow-up question generation (FQG) serves as a critical mechanism for improving information-seeking dialogues~\cite{hu-et-al-2024-followupsurvey}. Early work focused on clarification questions for resolving ambiguity~\cite{rahmani-etal-2023-survey, rao-daume-iii-2018-learning, gatto-etal-2025-follow}. However, for proactive systems targeting laypersons, clarification alone proves insufficient; instead, systems must provide follow-up questions alongside answers to guide a systematic exploration of complex topics~\cite {butala-etal-2024-promise}. 
The FOLLOWUPQG~\cite{meng-etal-2023-followupqg} dataset provides over 3K real-world (question, answer, follow-up question) triples from Reddit forums, offering layperson-friendly follow-ups but limiting the scope to single-turn dialogues. The TREC CAsT 2022 dataset~\cite{owoicho-et-al-2022-treccast} includes multi-turn dialogues with Wikipedia-grounded follow-up questions but remains general-domain without specialization for closed domains. These works demonstrate FQG's potential but reveal critical gaps, such as the lack of multi-turns, domain-specific proactive dialogue systems, and specialized FQG resources for complex information domains.

\paragraph{\textbf{Legal Domain Datasets and Systems.}}
The legal NLP landscape includes large-scale datasets and benchmarks like Pile of Law~\cite{henderson-et-al-2022-pile-of-law} and LegalBench~\cite{guha2023legalbench}. Question-answering (QA) resources encompass CUAD~\cite{hendrycks2021cuad}, PrivacyQA~\cite{ravichander-etal-2019-question}, and LegalQA~\cite{askari-et-al-2024-legalqa}. Recent conversational legal datasets include ConvLegal~\cite{liu-et-al-2025} for case retrieval dialogues and LegalConv~\cite{askari2023closer} with 360k layperson questions and 780k lawyer responses. However, LegalConv does not capture actual turn-taking dialogues.
LegalBot~\cite{john2017legalbot} provides short simulated layperson-lawyer exchanges extracted from law textbooks, averaging only two turns per dialogue. Existing research in conversational legal document access includes LexRAG~\cite{li2025lexrag} for Chinese legal statutes and Chat-EUR-Lex~\cite{cherubini2024improving} for EU regulations. However, these systems cannot engage in a proactive dialogue or systematically guide laypersons through complex legal topics. Critically, existing legal conversational datasets predominantly address legal advice and lack follow-up questions designed for structured topic exploration. 

We identify and address two critical gaps: (1) the absence of proactive multi-turn dialogue datasets in specialized domains, and (2) the lack of systematic approaches for realizing proactive dialogue systems that will guide laypersons toward specific information through structured follow-up question generation.
\section{The EUDial Dataset}
\label{sec:datasetmodel}

\paragraph{\textbf{Document model and task definition.}} In the following, we detail the construction of dialogue datasets from AskEP\footnote{\url{https://epthinktank.eu/author/epanswers}, accessed on 29th August 2025} blogs curated by the Citizens’ Enquiries Unit (AskEP) within the European Parliamentary Research Service (EPRS). 
Each blog entry is represented as a document $d_i = \langle q_i, a_i \rangle$, where $q_i$ is the citizen's question and $a_i$ is the expert response decomposed into sections $a_i = [s^i_{1}, \ldots, s^i_{j}]$. Each section $s^i_{j}$ is modeled as a pair $\langle h^i_{j}, c^i_{j} \rangle$, with header $h^i_{j}$ and content $c^i_{j}$. Documents also include metadata such as publication date, hyperlinks, and thematic tags.

We view a document $d_i$ as a single-turn dialogue, where a user query $q_i$ is followed by an answer $a_i$. The task is to transform this into a multi-turn dialogue $T_i = \{ t^i_{1}, \ldots, t^i_{j} \}$, where each dialogue turn is defined as:
\begin{equation}
\label{eq:conversational-dataset}
t^{i}_{j} = \langle u^{i}_{j}, r^{i}_{j}, f^{i}_{j} \rangle,
\end{equation}
with $u^{i}_{j}$ being the user utterance, which at each turn $t^{i}_{j}$ may be (i) the initial user query, (ii) a system-proposed follow-up query when acknowledged by the user, or (iii) a new user query posed when the system-proposed follow-up query is not acknowledged. The system reply $r^{i}_{j}$ is grounded in $c^i_{j}$, and $f^{i}_{j}$ is a proactive follow-up query guiding further exploration. Algorithm~\ref{alg:dataset} (Appendix~\ref{appendix:datasetalgorithm}) describes the procedure for converting a single-turn QA pair into a multi-turn dialogue. Appendix~\ref{appendix:qapair} illustrates an example QA pair collected from AskEP and its corresponding structured representation. The resulting dataset $T = \{T_1,\ldots,T_n\}$ thus enables the evaluation of dialogue systems that respond and initiate systematic knowledge acquisition.

\paragraph{\textbf{Dataset construction.}} The EUDial dataset consists of multi-turn dialogues with section-level references and supporting legal documents, systematically transformed from AskEP blogs. We collected 204 blogs published between January 2014 and July 2025, each addressing a specific citizen enquiry (question) through multi-sectioned expert responses (answer). Blogs are scraped into QA pairs while preserving their section structure and metadata. Since citizen questions often appear ambiguously (e.g., as titles or introductory paragraphs), GPT-4o-mini\footnote{\url{https://platform.openai.com/docs/models/gpt-4o-mini}, accessed on 2nd September 2025} is prompted to generate standardized question formulations (see also Appendix~\ref{appendix:qg}). For each QA pair, multi-turn dialogues are then generated using GPT-4o-mini by mapping sections to dialogue turns through a controlled prompt (see also Appendix~\ref{appendix:dialoguegeneration}) and following Algorithm~\ref{alg:dataset} (Appendix~\ref{appendix:datasetalgorithm}). Generation of the full dataset required approximately 45 minutes on a standard personal computer (16 GB RAM, ARM-based processor), incurring a total cost of \$0.11. Importantly, hyperlinks embedded in response sections are preserved as references to legal documents, ensuring that each dialogue turn provides verifiable legal grounding.

\begin{table}[!t]
\centering
\caption{Comparison of EUDial with existing conversational legal and follow-up question datasets. A dialogue is proactive when the system not only answers but also generates follow-up questions to guide exploration, while reactive systems only respond. One dialogue turn consists of a user utterance and a system reply (plus a follow-up in proactive). \emph{Attribution} indicates response grounding in external sources.}
\resizebox{\linewidth}{!}{%
\begin{tabular}{lcccc}
\toprule
\textbf{Feature} & \textbf{EUDial} & \textbf{Legalbot} & \textbf{FOLLOWUPQG} & \textbf{ProMISe} \\
\midrule
Domain & EU law/policy & Law texbooks & Open-domain & Open-domain \\
Dialogue type & Multi-turn, proactive & Multi-turn, reactive & Single-turn, proactive & Multi-turn, proactive \\
\# dialogues & 204 & 1200 & 3790 & 1025 \\
\# dialogue turns & 880 & 2400 & 3790 & 4453\\
Avg \# dialogue turns & 4.3 & 2 & 1 & 4.3\\
Attribution & Yes & No & No & No \\
\bottomrule
\end{tabular}%
}
\label{tab:dataset-comparison}
\end{table}

As shown in Table~\ref{tab:dataset-comparison}, existing resources like Legalbot focus primarily on reactive dialogues (legal advice) without proactive follow-up mechanisms. At the same time, FOLLOWUPQG provides open-domain single-turn follow-up triples. ProMISe represents a proactive, multi-turn dialogue dataset in an open domain but operates in a two-player simulation without grounding in authoritative sources. The EUDial dataset is constructed from human-written AskEP blogs, where each conversation is explicitly grounded to the original blog content and referenced legal documents, thus acting as a benchmark dataset for proactive dialogue systems in the legal domain.

\paragraph{\textbf{Dataset Assumption.}} The dialogue generation process operates under assumptions that define the scope of conversational behaviors captured in EUDial. First, follow-up questions are generated sequentially for each subsequent section without multiple questions targeting the same section content. Second, we assume users exhibit consistent engagement by always responding affirmatively to explore presented information, creating a linear progression through all available sections. As a result, the dataset does not capture alternative conversational paths such as users declining information, requesting different topics mid-conversation, or terminating dialogue before entire topic coverage. These assumptions enable the creation of a systematic dataset, but they represent areas for future extension to more complex user behaviors.

\paragraph{\textbf{Dataset Characteristics.}} 
The AskEP blog entries that seed the EUDial dataset exhibit structural properties making conversational modeling both necessary and beneficial. Each answer has multi-sections (3.9 on average; maximum of 9) with substantial section-level content (mean of 155 words). Questions are comparatively short (mean length is around 18 words) relative to answers, yielding a very low question-to-answer ratio ($\approx$0.04) at the QA-pair level, which indicates that user queries provide narrow entry points into broader content. Readability is challenging for lay users (Flesch Reading Ease =26.2 on average), indicating a “difficult” level. These characteristics highlight the need for a dialogue dataset that can (i) decompose long, multi-section answers into navigable units, and (ii) generate accessible follow-up questions to guide users through otherwise inaccessible legal material. 

The resulting EUDial dialogue dataset comprises 204 dialogues, with each dialogue averaging 4.3 turns (ranging from 2 to 18 turns), and alternating roles between citizens (layperson) and EPRS (system).  Importantly, each system turn is grounded in the underlying legal resources used to construct the answer. Moreover, every system response not only provides an answer to the user’s question but also includes a follow-up question that introduces the next section of the legal content, thereby guiding the dialogue forward. Thus, EUDial acts as a benchmarck dataset for proactive dialogue systems. Further dataset insights for the collected AskEP blogs and constructed EUDial dialogue dataset are provided in Appendix~\ref{appendix:datasetinsights}.
\section{The LexGuide Framework}
\label{sec:dialogue-framework}

\begin{figure}[t]
  \centering
  \includegraphics[width=\linewidth]{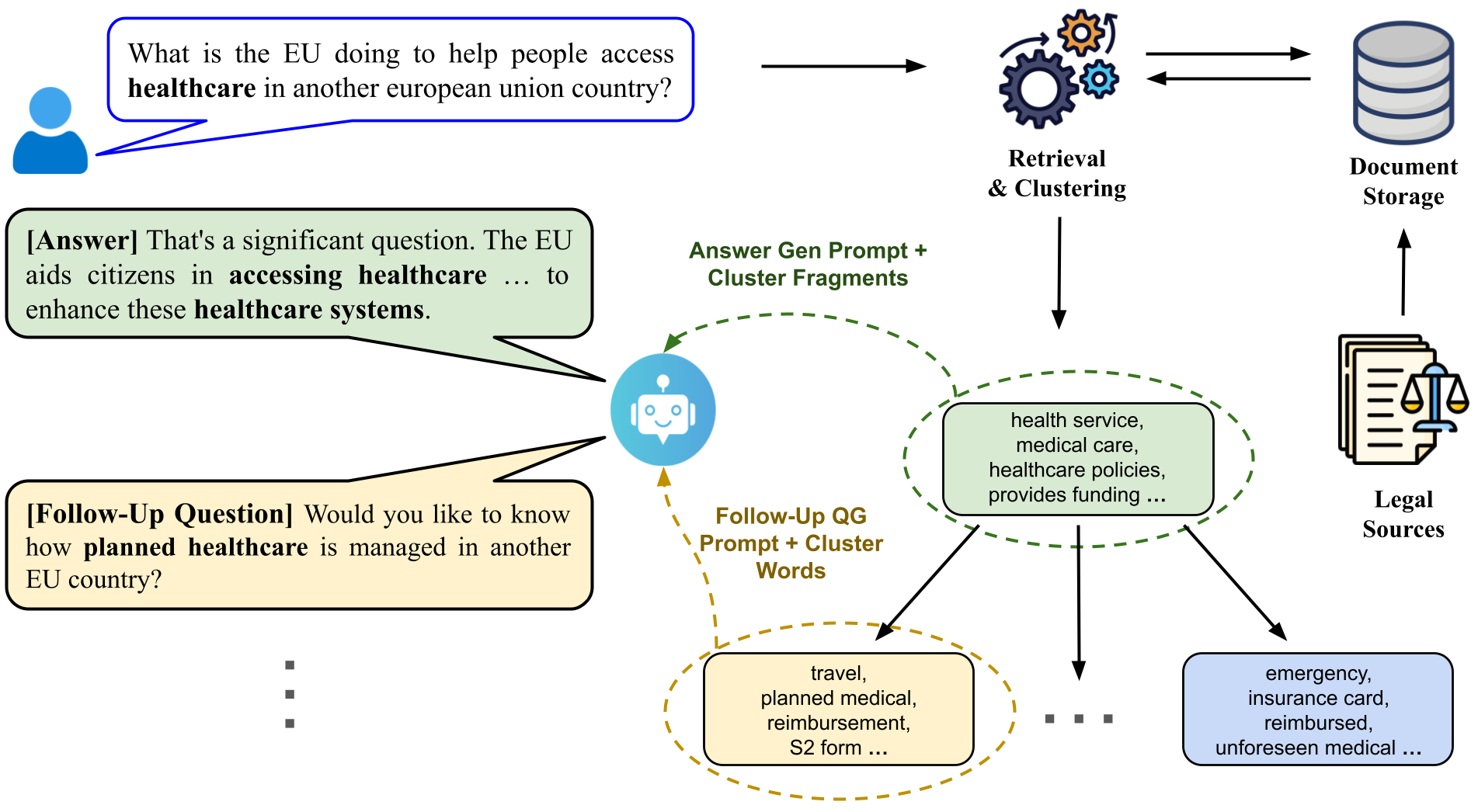}
  \caption{Overview of the LexGuide framework, illustrated with an example of the first turn in a multi-turn dialogue. The framework processes initial user requests, constructs topic tree representations, and leverages this structure to generate an answer and a follow-up question for guiding the user.}
  \label{fig:lexguide}
\end{figure}

EUDial dialogue dataset (Section~\ref{sec:datasetmodel}) serves as a benchmark dataset for evaluating proactive multi-turn dialogue systems. Building on EUDial, we propose the LexGuide framework (see Figure~\ref{fig:lexguide}), a proactive dialogue framework integrating Retrieval-Augmented Generation (RAG) with hierarchical topic organization and system-driven follow-up question generation to guide users through legal information structurally. For a dialogue $d_i$ with turns $t^{i}_{1}, t^{i}_{2}, \ldots$, at each dialogue turn $t^{i}_j$ with user utterance $u^{i}_{j}$, the framework executes hierarchical topic organization for structuring retrieved information, and uses RAG for response and follow-up question generation.

\paragraph{\textbf{Corpus preparation.}}  
Let $X = \{x_1, \ldots, x_m\}$ denote a legal corpus of $m$ documents, constructed from legal sources referenced in the EUDial dataset (refer to Figure~\ref{fig:sourcedistribution} in Appendix~\ref{appendix:datasetinsights} for the distribution of legal sources). Each document $x_i$ is modeled as an ordered sequence of fragments $P_i = [p^{i}_{1}, p^{i}_{2}, \ldots, p^{i}_{l}]$, where each fragment is a sequence of tokens, in most cases a sentence or a sequence of sentences. These fragments are embedded into a vector space and indexed for retrieval.

\paragraph{\textbf{Hierarchical organization.}}  
On the initial dialogue turn $t^{i}_1$, the system retrieves the top-$k$ fragments $R$ that are most relevant to the user utterance $u^{i}_{1}$, i.e., the user's initial query. To reduce fragment redundancy, retrieval employs diversity-aware \emph{Maximum Marginal Relevance} (MMR)~\cite{carbonell1998use}.
The retrieved fragments $R$ are clustered using BERTopic~\cite{grootendorst2022bertopic}, which leverages fragment embeddings followed by density-based clustering to identify clusters. To capture structured knowledge representation, hierarchical clustering organizes clusters into parent-child relationships. Each cluster represents a topic, and term frequency-inverse document frequency (tf-idf) is used to extract topic representations. 
This creates a topic tree $G_{i} = (V, E)$ where $V = \{g_0, g_1, \ldots, g_H\}$ represents topic nodes, and $E$ specifies relationship between nodes. The root node $g_0$ encompasses the broadest thematic coverage, while leaf nodes represent subtopics. Each intermediate level provides progressively refined semantic granularity, enabling users to explore information at their preferred level of detail. 

Each topic node $g \in V$ is formally characterized by a triple $\langle W_g, C_g, F_g \rangle$, where $W_g = \{w^g_1, w^g_2, \ldots, w^g_{l}\}$ is an ordered set of $l$ representative words ranked by topic relevance score, $C_g \in \mathbb{R}^d$ is the topic centroid computed as the mean embedding of associated fragments, and $F_g \subseteq R$ is the subset of fragments assigned to topic $g$. This hierarchical topic organization transforms the initially flat retrieved fragment set into a structured exploration space, where each level of the tree $G$ offers distinct semantic granularity for systematic dialogue navigation.

\paragraph{\textbf{Dialogue navigation and follow-up question generation.}}
The navigation state at turn $t^{i}_{j}$ is represented as
\begin{equation}
\label{eq:state}
S^{i}_j = (Y^{i}_{j}, L^{i}_{j}, Z^{i}_{j}, B^{i}_{j}, K^{i}_{j}, H^{i}_{j}),    
\end{equation}
where $Y^{i}_{j}$ is the set of visited nodes, $L^{i}_{j}$ is the current node, $Z^{i}_{j}$ are unexplored children of $L^{i}_{j}$, $B^{i}_{j}$ is the navigation strategy (Breadth-First Search (BFS), Depth-First Search (DFS), or user-driven), $K^{i}_{j}$ is the path of visited nodes, and $H^{i}_{j}$ is the dialogue history composed of user utterance $u^{i}_{j}$, response $r^{i}_{j}$, and follow-up query $f^{i}_{j}$. 

At each turn $t^{i}_{j}$, dialogue navigation is determined by operations, i.e., \textit{descend} into a child cluster for finer exploration, \textit{lateral} movement across sibling clusters for related themes, \textit{ascend} to a parent cluster for broader context, \textit{jump} to any cluster in the hierarchy for direct access, or \textit{backtrack} using path to previously visited states for re-exploration. Due to navigation state (as shown in Equation~\ref{eq:state}), users are provided flexiblity to return to the previous turn ($t^{i}_{j-1}$), revisit an earlier turn ($t^{i}_{j-n}$), returning to the point where a desired topic was active, or reopen a past discussion point to explore alternative trajectories.

The framework employs navigation strategies $B^{i}_{j}$ based on dialogue context and topic characteristics: (1) BFS explores sibling clusters at the current level before descending, i.e., \textit{lateral} operation, ensuring comprehensive coverage of related themes, (2) DFS follows topic hierarchies to a maximum depth, enabling detailed exploration of a specific topic, i.e., \textit{descend} operation, and (3) a user-driven strategy adapts to explicit user navigation requests, i.e., \textit{ascend}, \textit{jump}, or \textit{backtrack} operations. When a user acknowledges a system-proposed follow-up ($u^{i}_{j+1} = f^{i}_{j}$), the framework continues traversal within $G$ using the selected strategy, with the default being BFS for diverse topical coverage. If the user does not acknowledge a follow-up and instead poses a new query $q'$, the framework first checks whether $q'$ can be addressed by revisiting nodes that have already been explored ($Y^{i}_{j}$) or by descending into unexplored children of the current node ($Z^{i}_{j}$). If $q'$ cannot be resolved through these options, the framework computes coverage relative to the active tree $G$ as $Cov(G,q')=\max_{g\in G}\textit{sim}(q',C_{g})$. Considering $\tau$ as the topic coverage threshold, if $Cov(G,q') \geq \tau$, the framework directly navigates to the best-matching cluster. If $Cov(G,q') < \tau$, the user is prompted to clarify whether to shift to a new topic, in which case a new tree $G'$ is constructed. Depending on the user’s response, the framework flexibly applies these operations while RAG filters fragments associated with the active node $L^{i}_{j}$. These filtered fragments are concatenated with $u^{i}_{j}$ and prompted to the LLM for a response generation $r^{i}_{j}$. Each turn is thus a triple $\langle u^{i}_{j}, r^{i}_{j}, f^{i}_{j} \rangle$, where $f^{i}_{j}$ denotes the generated follow-up, consistent with the dialogue model defined in Eq.~\ref{eq:conversational-dataset}.

The dialogue navigation procedure identifies the next topic \(g_{\text{next}}\) to explore and follow-up question is generated using \(W_{\text{next}}\) and dialogue history $H^{i}_{j}$,
\begin{equation}
\label{eq:follow-up}
    f^{i}_{j} = \text{LLM}\big(\text{prompt}(W_{\text{next}}.H^{i}_{j})),
\end{equation}
where \(W_{\text{next}}=\{w_1,\ldots,w_l\}\) denotes the top-\(l\) topic representative words for the next topic $g_{\text{next}}$. A dialogue terminates under one of three conditions: (i) complete topic coverage when all nodes in the topic tree have been visited, (ii) explicit user satisfaction signals detected, or (iii) dialogue abandonment indicated by user inactivity or explicit termination requests. Algorithm~\ref{alg:lexguide-framework} (Appendix~\ref{appendix:dialeualg}) formalizes the procedure, showing how the framework gradually narrows from broad responses to fine-grained exploration while allowing flexible pivots.
\section{Experimental Setup}
\label{sec:experiment-setup}
For assessing the effectiveness of the proposed LexGuide framework, we design an experiment that evaluates both response generation quality and dialogue navigation performance. The evaluation is grounded in a carefully constructed legal corpus derived from the EUDial dataset. Our setup emphasizes three aspects: (i) corpus preparation and preprocessing to ensure a representative evaluation resource, (ii) methodological comparisons against established baselines, and (iii) automated evaluation metrics to evaluate factual accuracy, completeness, legal relevance, navigational effectiveness, and follow-up quality. This integrated approach enables us to examine not only whether LexGuide generates relevant answers, but also whether it systematically guides users through legal topics in a proactive and interpretable manner.

We construct the experimental corpus from the EUDial metadata by considering 2450 hyperlinks (legal source URLs). After URL normalization and de-duplication, 2036 unique URLs remain. Removing 181 social-media endpoints yields 1855 processable URLs.
Of these, 1329 are web pages, 91 are PDF documents, and 435 return HTTP errors. The final corpus used in experiments comprises 1420 processable URLs. The corpus construction process begins with content extraction using unstructured.io\footnote {\url{https://unstructured.io}, accessed on 7th September 2025} for web pages and DocLing~\cite{Docling} for PDFs. Fragments obtained while content extraction are embedded using \textit{nlpaueb/bert-base-uncased-eurlex}~\cite{chalkidis-etal-2020-legal}, and indexed in FAISS~\cite{johnson-et-al-2019-billion} for retrieval. Retrieval uses MMR with diversity hyperparameter $\lambda=0.6$ for broader topic coverage. For text generation, we evaluate GPT-4o-mini for efficient processing with strong reasoning capabilities, llama-3.1-8b-instant\footnote{\url{https://console.groq.com/docs/model/llama-3.1-8b-instant}, accessed on 8th September 2025} and mixtral$-$8x7B model\footnote{\url{https://ollama.com/library/mixtral:8x7b} (accessed on 12th September 2025)} for open-source performance, and gemma2-2b\footnote{\url{https://ollama.com/library/gemma2:2b}, accessed on 12th September 2025} for resource-constrained deployment scenarios. Temperature settings are maintained at 0.3 across all models for balanced creativity and consistency in response generation and follow-up question formulation.

Topic modeling at the first dialogue turn $t^{i}_{1}$ utilizes BERTopic with UMAP~\cite{McInnes2018} dimensionality reduction, HDBSCAN~\cite{Campello2013} for density-based clustering, and tf-idf to extract representative keywords over cluster fragments. To obtain a hierarchy, hierarchical clustering employs agglomerative clustering with \emph{average} linkage. This procedure yields a topic tree $G=(V,E)$ whose depth is not fixed a priori; in the current EUDial dataset, dialogues yield shallow hierarchies of depth two, reflecting that citizen questions tend to anchor a dominant theme with a small number of focused subtopics. In other domains or dialogue styles, the same procedure may yield deeper trees without modification. Due to these characteristics of the EUDial dataset during the dialogue navigation in the LexGuide framework, the BFS navigation strategy is considered for lateral sibling navigation. 

For the initial user utterance of a dialogue, the system retrieves fragments using diversity-aware MMR with $\lambda=0.6$ and considering top-$k$ fragments with $k=500$. These retrieved fragments are used for hierarchical topic modeling, while immediate answering relies on a narrower subset of fragments from the node in focus ($k=10$). 
The follow-up generator consumes only the next node’s representative words with turn context (refer to Equation~\ref{eq:follow-up}). 
\section{Evaluation}
\label{sec:evaluation}

\paragraph{\textbf{Evaluation Methodology and Metrics.}} We evaluate the proposed LexGuide framework using the EUDial dataset (comprised of 204 multi-turn dialogues), focusing on response generation and dialogue navigation. For each dialogue, the LexGuide framework initializes the system with the first user question, constructs the hierarchical topic tree $G$, and executes the navigation-driven policy that generates answers and proposes follow-ups until the termination criteria are met, i.e., all sub-topics are explored. We compare against baselines: (i) \textit{RAG-Basic}, a single-turn RAG without MMR or clustering; (ii) \textit{RAG-MMR}, incorporating diversity-aware retrieval without clustering; and (iii) \textit{ConvRAG}, a conversational RAG generating follow-ups without topic modeling or navigation state tracking. All baselines use the same FAISS index and embedding model (\texttt{nlpaueb/bert-base-uncased-eurlex}).

The \textit{answer quality} is evaluated along: (i) \textit{groundedness}, measured as the proportion of sentences overlapping with retrieved fragments; (ii) \textit{completeness}, computed as ROUGE-L recall against gold responses; (iii) \textit{relevance}, measured with BERTScore~\cite{bertscore} to capture semantic similarity between generated and gold responses; and (iv) \textit{layperson readability}, measured using the Flesch Reading Ease (FRE)~\cite{flesch1948new} score estimating textual accessibility for non-experts. Including FRE enables us to complement legal correctness with accessibility, thereby capturing whether system responses are not only accurate but also comprehensible to lay audiences.

\textit{Follow-up quality} is assessed in terms of (i) \textit{relevance}, measured as similarity between generated follow-ups and gold expert follow-ups; (ii) \textit{diversity}, average pairwise distance between follow-ups generated within dialogues; and (iii) \textit{temporal consistency}, measuring conversational flow coherence. \textit{Topic coverage} evaluates the proportion of gold-relevant sub-topics discussed during the dialogue.

\begin{table}[!ht]
\centering
\caption{Comparative evaluation of response quality across baselines and LexGuide.}
\label{tab:response-quality}
\begin{tabular}{lccccc}
\toprule
\textbf{Method} & \textbf{Model} & 
\makecell{\textbf{Grounded-}\\\textbf{ness (\%)}} & 
\makecell{\textbf{Completeness}\\\textbf{(ROUGE-L)}} & 
\makecell{\textbf{Relevance}\\\textbf{(BERTScore)}} & 
\makecell{\textbf{Readability}\\\textbf{(FRE)}} \\
\midrule
RAG-Basic & LlaMa-3.1-8B & 96.4 & 0.40 & 0.72 & 96.1 \\
RAG-MMR   & LlaMa-3.1-8B & 96.4 & 0.40 & 0.72 & 96.1 \\
ConvRAG   & LlaMa-3.1-8B & 97.7 & \textbf{0.42} & 0.74 & 96.5 \\
\midrule
LexGuide & LlaMA-3.1-8B & 96.5 & 0.38 & 0.73 & 96.6 \\
LexGuide & GPT-4o-mini & \textbf{98.7} & 0.37 & \textbf{0.76} & 96.2 \\
LexGuide & Gemma2-2B & 92.9 & 0.37 & 0.72 & 92.7 \\
LexGuide & Mixtral$-$8x7B & 96.3 & 0.40 & 0.75 & \textbf{98.5} \\
\bottomrule
\end{tabular}
\end{table}

\paragraph{\textbf{Results on Answer Quality.}}  
As shown in Table~\ref{tab:response-quality}, LexGuide demonstrates competitive performance in answer quality metrics, with notable architecture-dependent variations. The GPT-4o-mini variant achieves the highest groundedness score (98.7\%) and relevance score (0.761) across all methods. Readability remains consistently high across all variants, indicating that the inclusion of layperson-oriented phrasing is preserved.

\paragraph{\textbf{Results on Navigation and Follow-up Quality.}}  
Topic coverage evaluation reveals LexGuide's novel navigation capabilities. Since baselines lack topic hierarchies, their topic coverage relies on content-based (86.6-87.6\%). LexGuide's topic-word-based navigation achieves 73.7\% topic coverage across all models, while content-based evaluation shows competitive performance (87.3-89.5\%), confirming that hierarchical exploration maintains topic relevance. LexGuide’s topic-word-driven design promotes substantial diversity (0.204-0.224) in follow-up questions compared to baseline methods (0.201-0.205), helping prevent redundancy in dialogues. The generated follow-ups remain semantically relevant and contextually linked to prior turns, thereby supporting coherent exploration. 

Thus, LexGuide structured topic navigation maintains high-quality dialogue and framework consistency across model architectures (identical 73.7\% topic coverage) demonstrates robust hierarchical navigation. The slight reduction in completeness reflects LexGuide's design choice of prioritizing depth and relevance over exhaustive coverage, aligning with guided legal exploration rather than comprehensive but potentially overwhelming responses.


\section{Conclusion and Future Work}
\label{sec:conclusion}

This paper addresses the gap between the availability of legal information and citizen comprehension through EUDial, a benchmark dataset comprising 204 proactive multi-turn dialogues from European Parliamentary Research Service blogs, and LexGuide, a hierarchical topic-guided dialogue framework for proactive legal exploration. Empirical results demonstrate that LexGuide consistently maintains strong factual groundedness (98.7\% with GPT-4o-mini), semantic relevance, and readability, thereby addressing both legal accuracy and accessibility for lay audiences. The topic coverage evaluation reveals that while topic-word-based coverage (73.7\%) demonstrates structured hierarchical navigation, content-based coverage (87.3-89.5\%) confirms comprehensive topic relevance, validating the framework's guided exploration approach. The evaluation results highlight a strategic design trade-off, where LexGuide's 2.7\% reduction in completeness reflects a prioritization of focused, relevant responses over exhaustive coverage, aligning to deliver accessible legal information. The framework's model-agnostic design and consistent hierarchical navigation across architectures demonstrate robust applicability. In our ongoing work, we are incorporating human studies with legal experts and lay users to validate factual faithfulness and navigational clarity beyond automatic metrics.

\bibliographystyle{vancouver}
\bibliography{custom}

\appendix
\section{Use of Large Language Models}
In this paper, we use large language models (ChatGPT-5 from OpenAI and Claude Sonnet 4 from Anthropic) strictly for editorial and technical assistance. The tools were employed for proofreading, refining language for clarity and academic tone, and providing minor coding support (e.g., boilerplate generation and suggestions for efficiency). All intellectual contributions, research design, analyses, and conclusions remain entirely those of the authors. AI-generated content was reviewed, edited, and validated before inclusion.

\section{Ethics and Data Statement}
\textit{AskEP} blogs consist of curated informational texts intended for public communication and education; they are not legally binding sources or official legal documents. Our dataset construction preserves the original sectioning, phrasing, and expert content while clearly distinguishing machine-generated elements, specifically the generated citizen questions and dialogue turns created through GPT-4o-mini. All original metadata, including publication dates, topic tags, and embedded hyperlinks to legal sources, are retained to support reproducibility, enable temporal analyses of policy evolution, and facilitate potential future dataset updates. The transformation process maintains attribution to source sections and legal documents, ensuring transparency in the provenance of information.

\section{Example QA Pair}
\label{appendix:qapair}
As described in Section~\ref{sec:datasetmodel}, each AskEP blog entry is modeled as a document comprising the citizen question, a multi-sectioned answer, and associated metadata. To illustrate, we present below a representative example of how such a blog entry appears after document modelling.

For a document on access to healthcare in another European Union (EU) country\footnote{\url{https://epthinktank.eu/2025/04/01/what-is-the-eu-doing-to-help-people-access-healthcare-in-another-eu-country}, accessed on 29th August 2025}, collected information from AskEP is as follows:
\begin{itemize}
    \item \textbf{question ($q_i$):} ``What is the EU doing to help people access healthcare in another EU country?''
    \item \textbf{answer ($a_i$):} 
    \begin{itemize}
        \item \texttt{Planned healthcare:} Under EU law, you have the right to travel to another EU country to receive medical care (consultation with a specialist, surgery \ldots
        \item \texttt{Unplanned healthcare (emergency):} Should you need unforeseen medical treatment while on holiday, a business trip or while studying in another EU country for a short period \ldots
    \end{itemize}
    \item \textbf{Metadata ($meta(d_i)$):} \{date = 2025-04-01, tags = [cross-border healthcare, public health], links = [``Planned healthcare’’ $\to$ EU law, S2 form, \ldots; ``Unplanned healthcare’’ $\to$ European Health Insurance Card; \ldots]\}
\end{itemize}

\begin{figure}[t]
  \centering
  \begin{subfigure}[t]{0.30\linewidth}
    \centering
    \includegraphics[width=\linewidth]{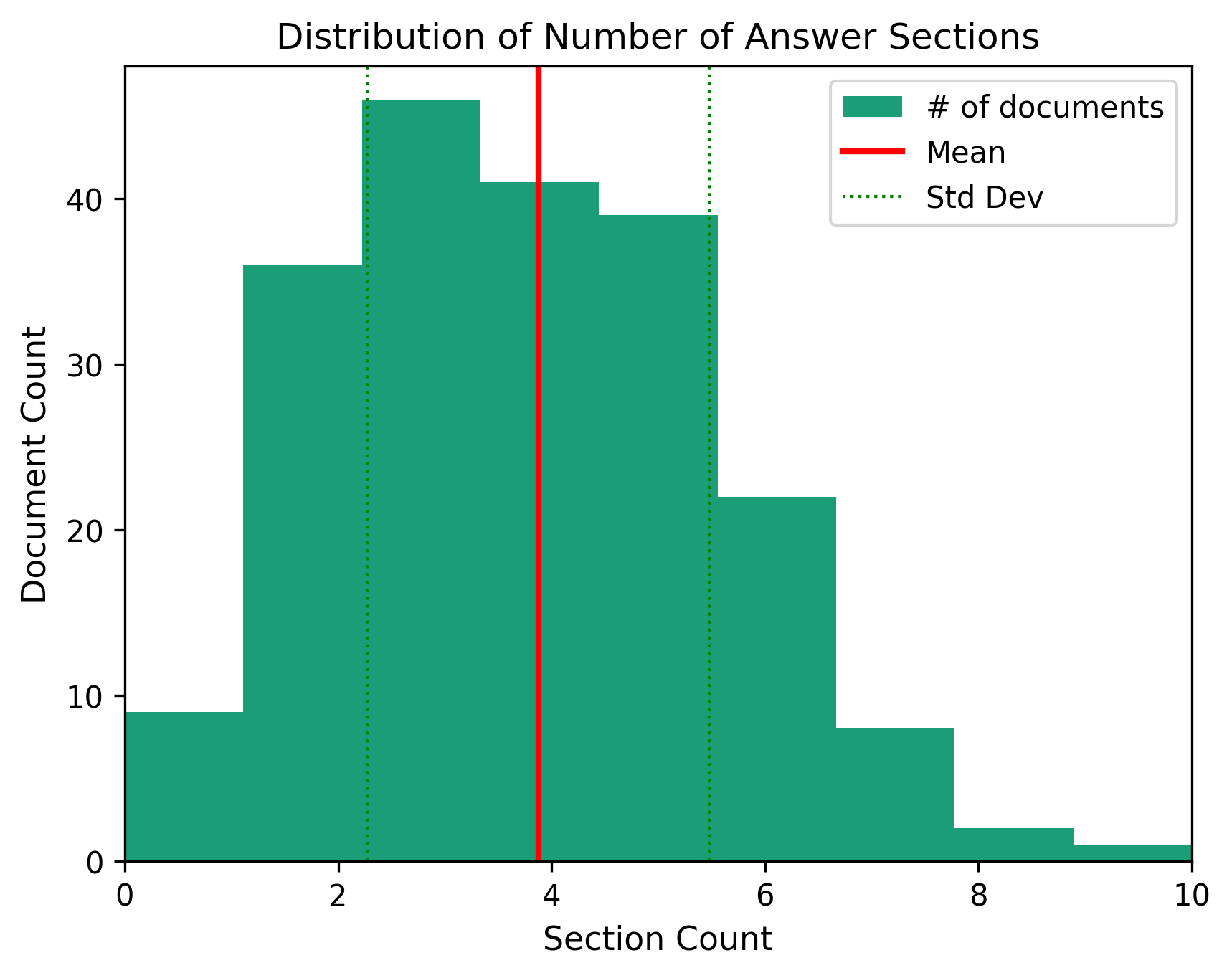}
    \caption{Sections per Answer}
    \label{fig:askep_section_counts}
  \end{subfigure}
  \hfill
  \begin{subfigure}[t]{0.30\linewidth}
    \centering
    \includegraphics[width=\linewidth]{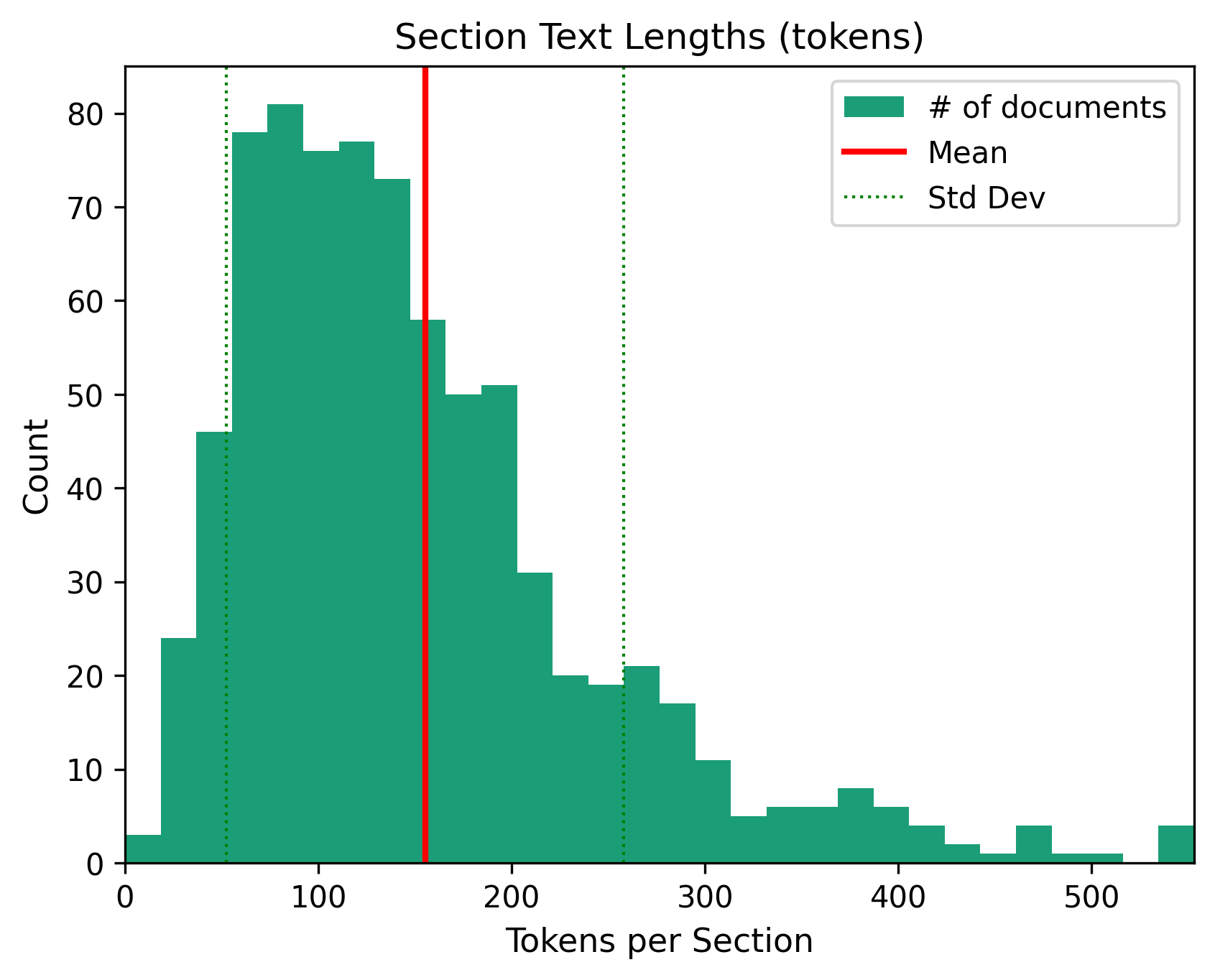}
    \caption{Section Lengths}
    \label{fig:askep_section_lengths}
  \end{subfigure}
  \hfill
  \begin{subfigure}[t]{0.30\linewidth}
    \centering
    \includegraphics[width=\linewidth]{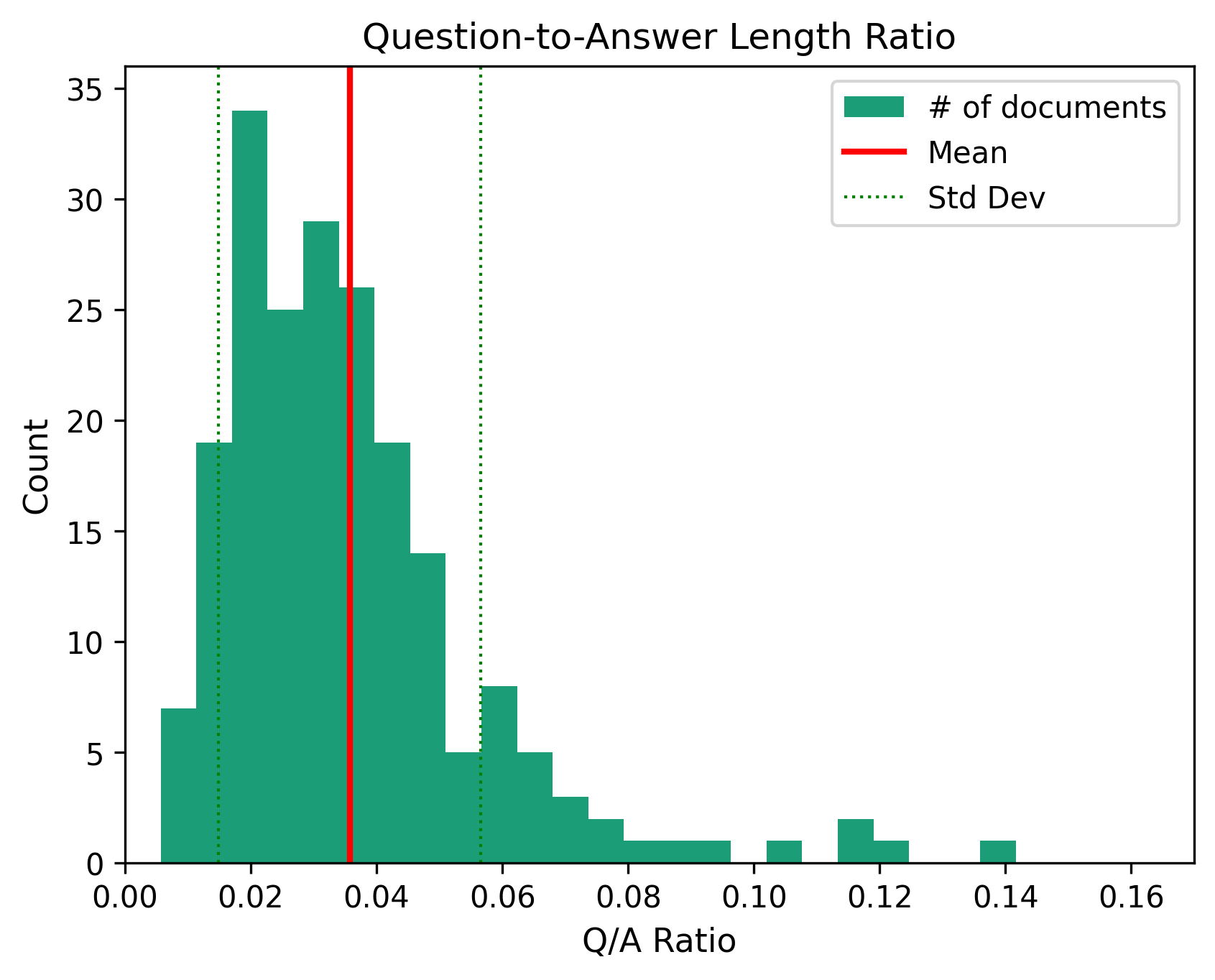}
    \caption{Q--A Ratio}
    \label{fig:askep_q2a}
  \end{subfigure}

  \vspace{1em}

  \begin{subfigure}[t]{0.30\linewidth}
    \centering
    \includegraphics[width=\linewidth]{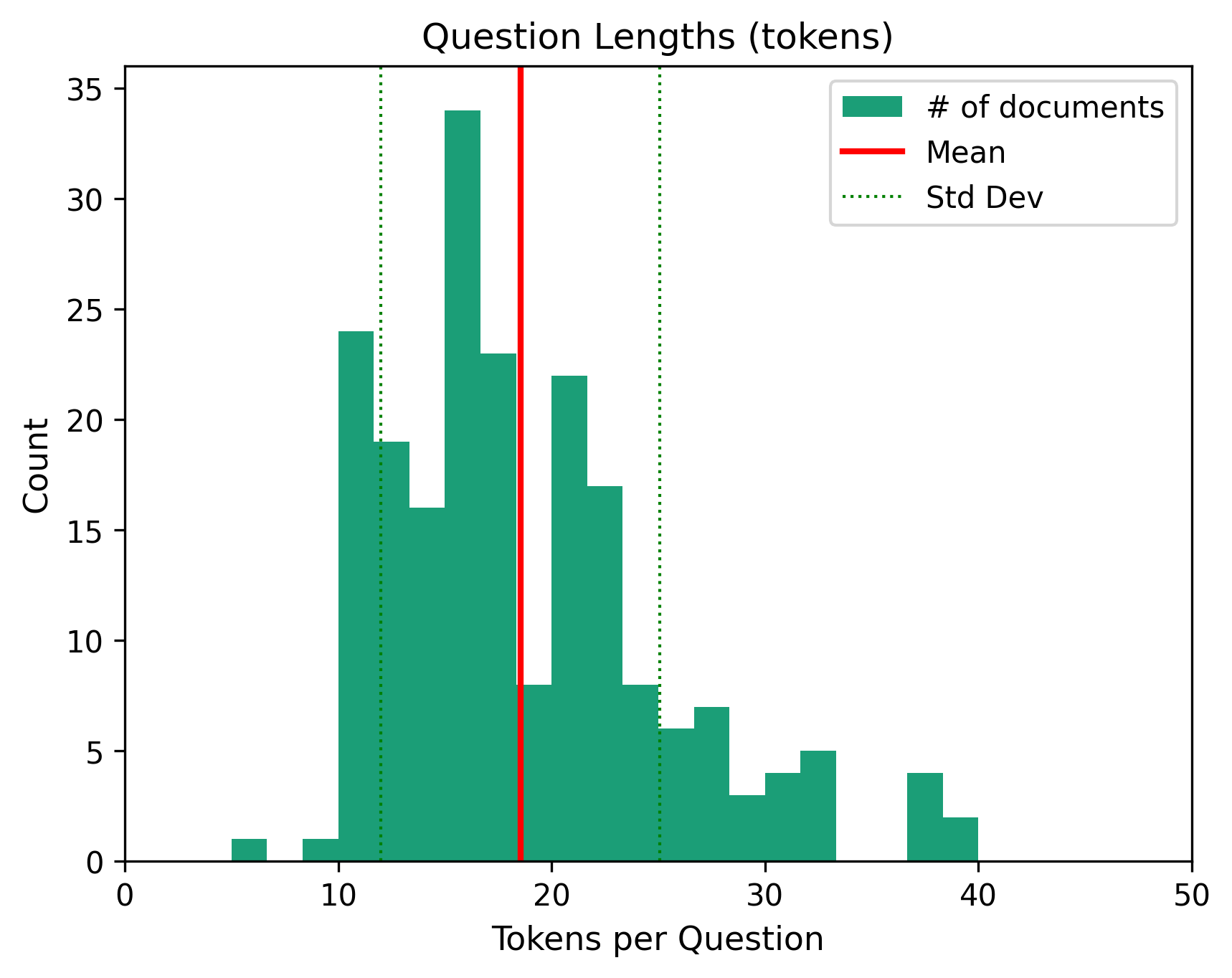}
    \caption{Question Lengths}
    \label{fig:askep_question_lengths}
  \end{subfigure}
  \hfill
  \begin{subfigure}[t]{0.30\linewidth}
    \centering
    \includegraphics[width=\linewidth]{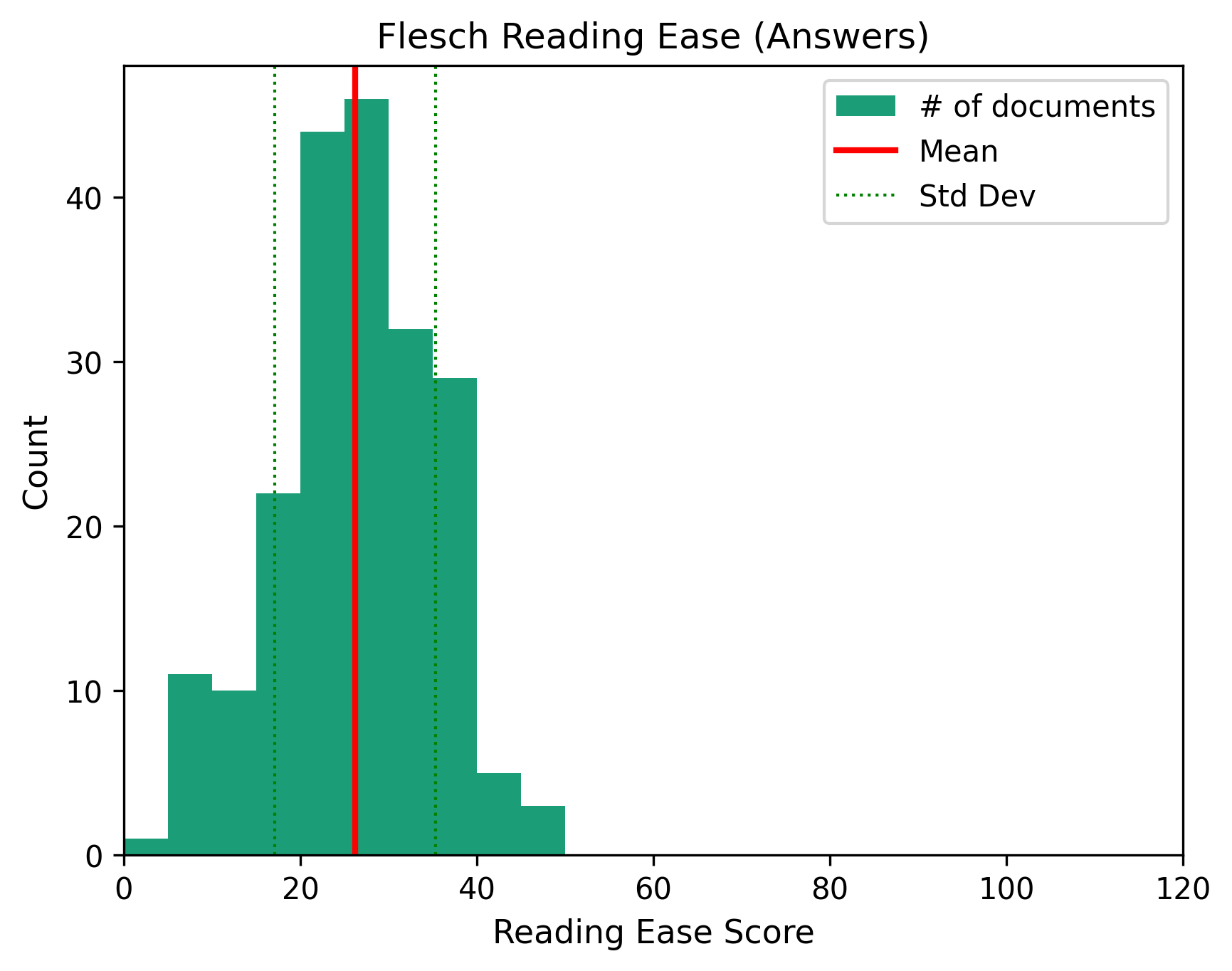}
    \caption{Readability (FRE)}
    \label{fig:askep_fre}
  \end{subfigure}
  \hfill
  \begin{subfigure}[t]{0.30\linewidth}
    \centering
    \includegraphics[width=\linewidth]{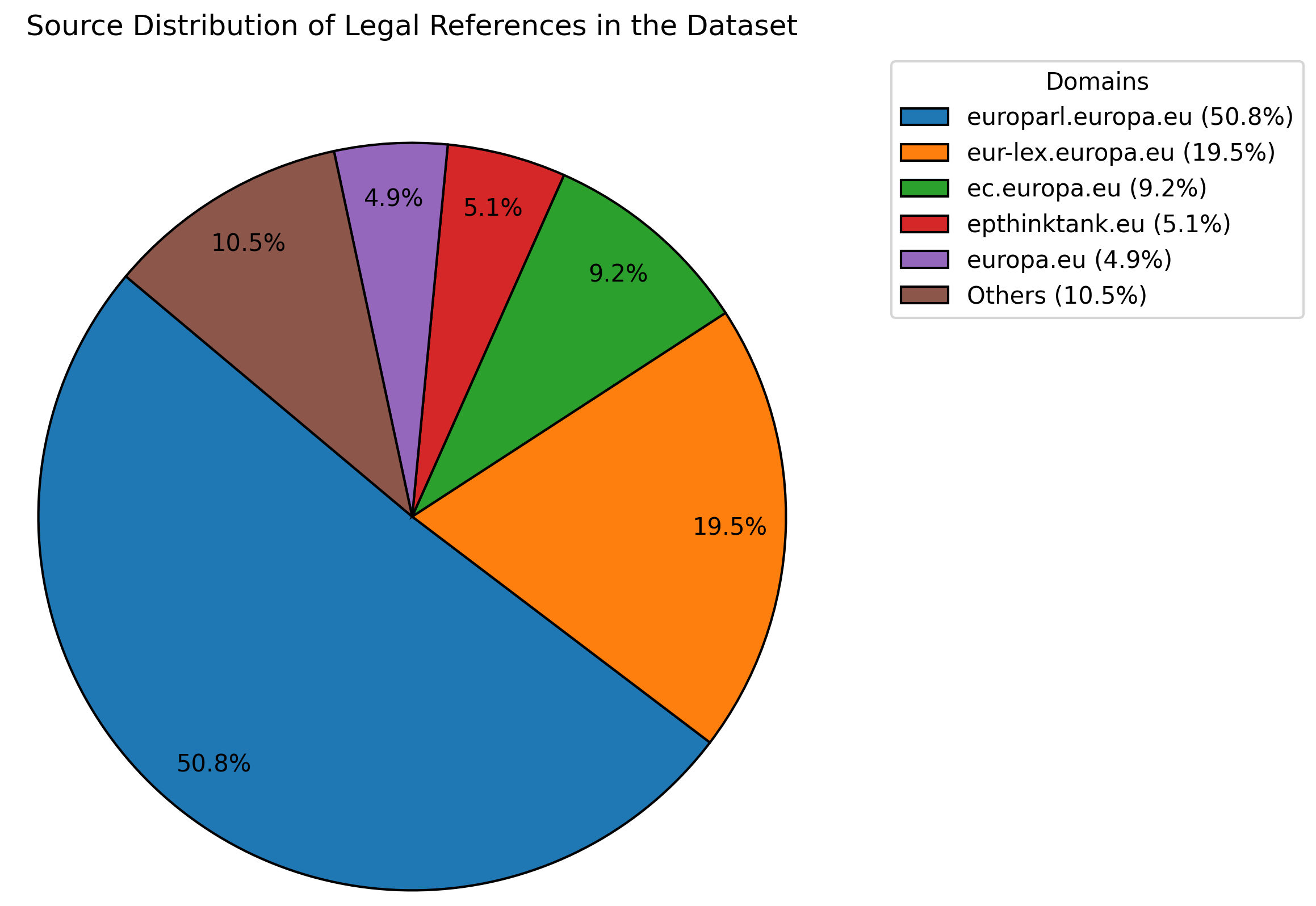}
    \caption{Legal resources}
    \label{fig:sourcedistribution}
  \end{subfigure}

  \caption{Distributional properties of \textit{AskEP} blog QA pairs: 
  (a) number of sections per answer, 
  (b) section lengths (tokens), 
  (c) question-to-answer length ratios, 
  (d) question lengths (tokens), 
  (e) readability scores (FRE), and 
  (f) distribution of legal resources.}
  \label{fig:askep_grid}
\end{figure}

\section{Dataset Insights}
\label{appendix:datasetinsights}

\paragraph{\textbf{AskEP blogs.}} Each expert-written answer from the AskEP blog entry comprises multiple sections (mean $=3.89$, maximum $=9$) (Figure~\ref{fig:askep_section_counts}). Answer sections vary widely, with an average of $155$ tokens and a maximum of $772$ words, indicating heterogeneity in information granularity (Figure~\ref{fig:askep_section_lengths}). The ratio between question length and total answer length averages $0.036$, confirming that questions are short entry points while answers introduce substantially more information (Figure~\ref{fig:askep_q2a}). User questions average 18 words, consistent with concise layperson queries (Figure~\ref{fig:askep_question_lengths}). Answers have poor readability for non-experts, with an average Flesch Reading Ease score of $26.2$ (Figure~\ref{fig:askep_fre}), suggesting that proactive dialogue support is necessary for accessibility. 

\begin{figure}[t]
  \centering
  \begin{subfigure}[t]{0.48\linewidth}
    \centering
    \includegraphics[width=\linewidth]{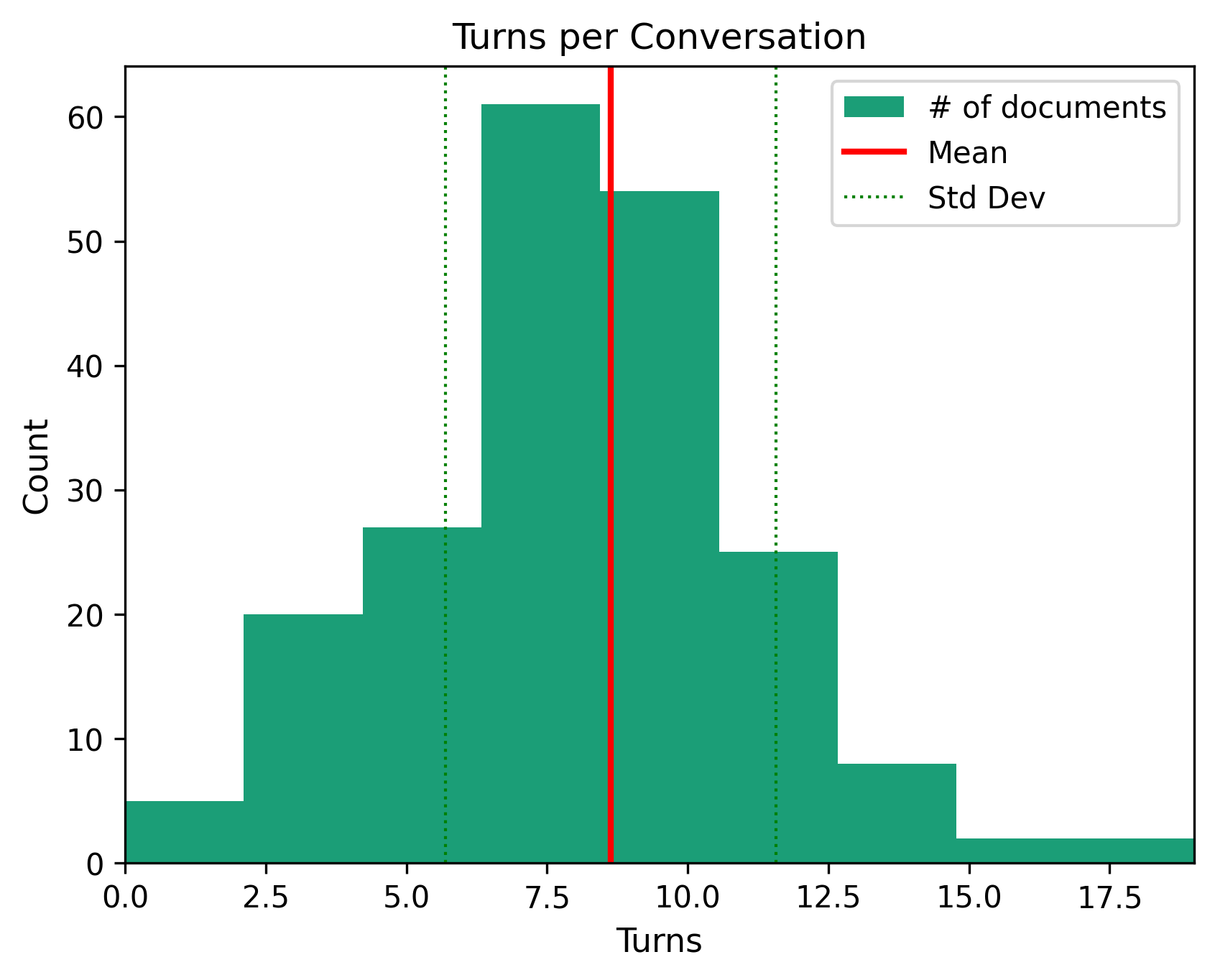}
    \caption{Turns per Conversation}
    \label{fig:conv_turns_per_conv}
  \end{subfigure}
  \hfill
  \begin{subfigure}[t]{0.48\linewidth}
    \centering
    \includegraphics[width=\linewidth]{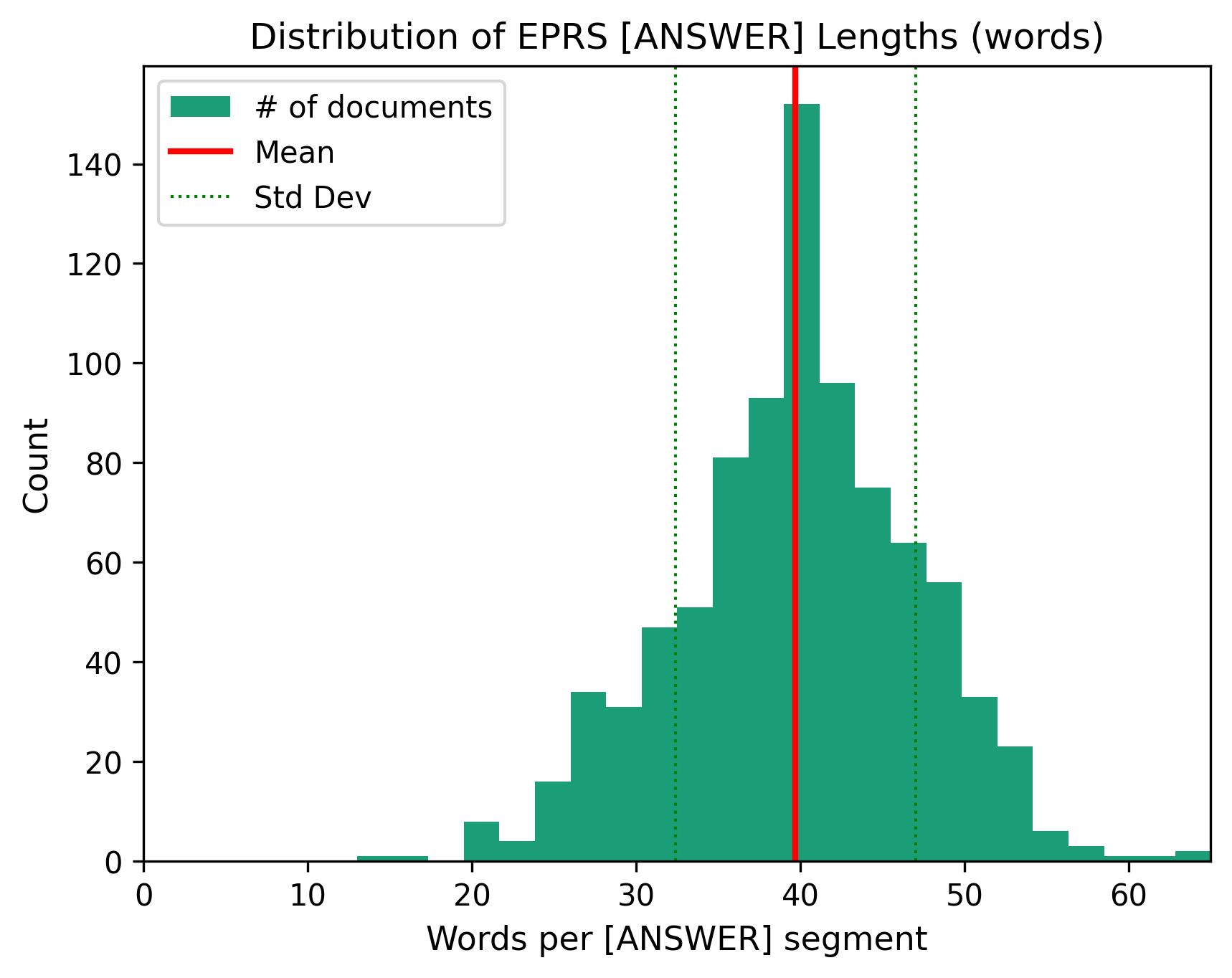}
    \caption{System Answer Length}
    \label{fig:conv_eprs_answer_lengths}
  \end{subfigure}

  \begin{subfigure}[t]{0.48\linewidth}
    \centering
    \includegraphics[width=\linewidth]{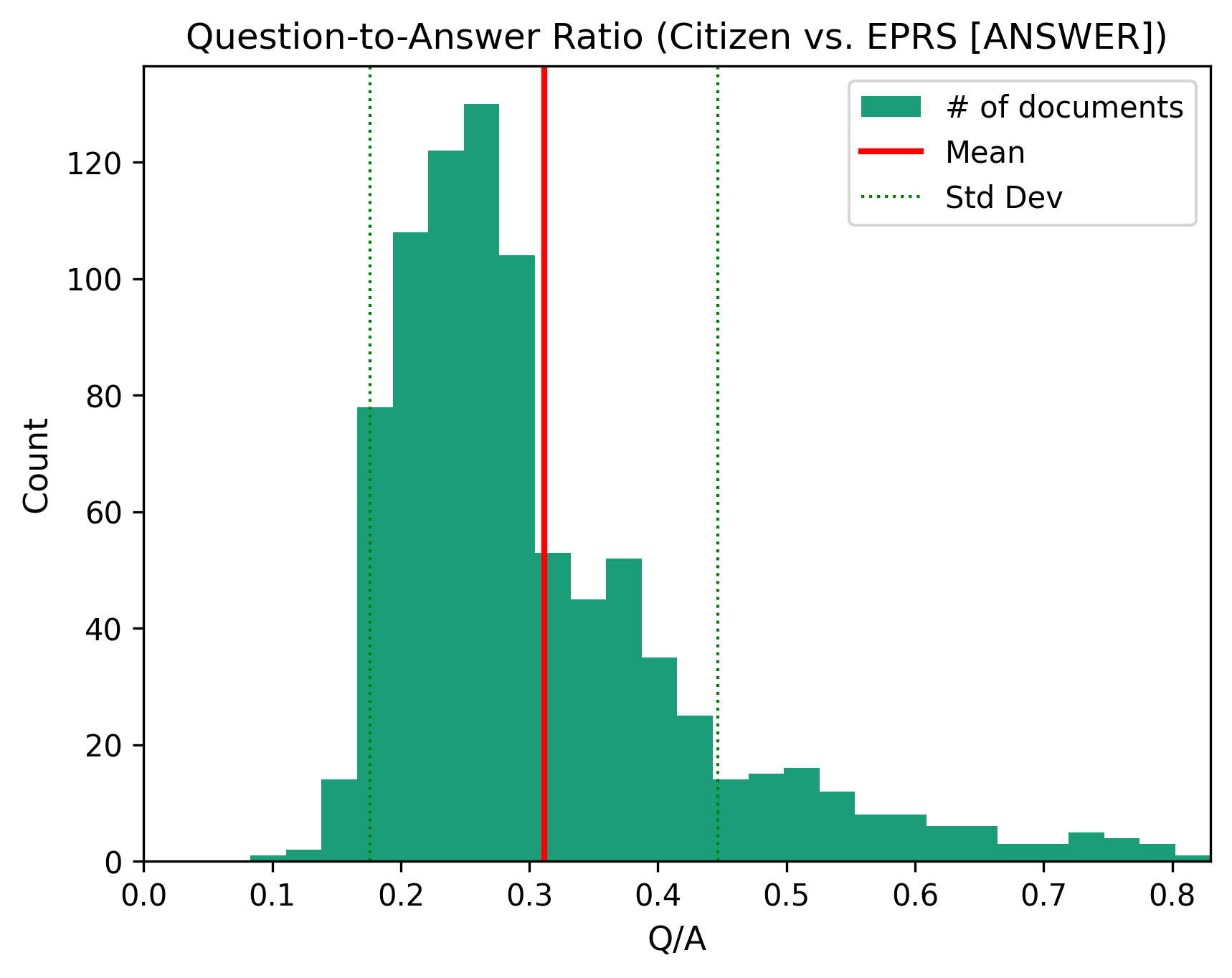}
    \caption{Q--A Ratio}
    \label{fig:conv_q2a}
  \end{subfigure}
  \hfill
  \begin{subfigure}[t]{0.48\linewidth}
    \centering
    \includegraphics[width=\linewidth]{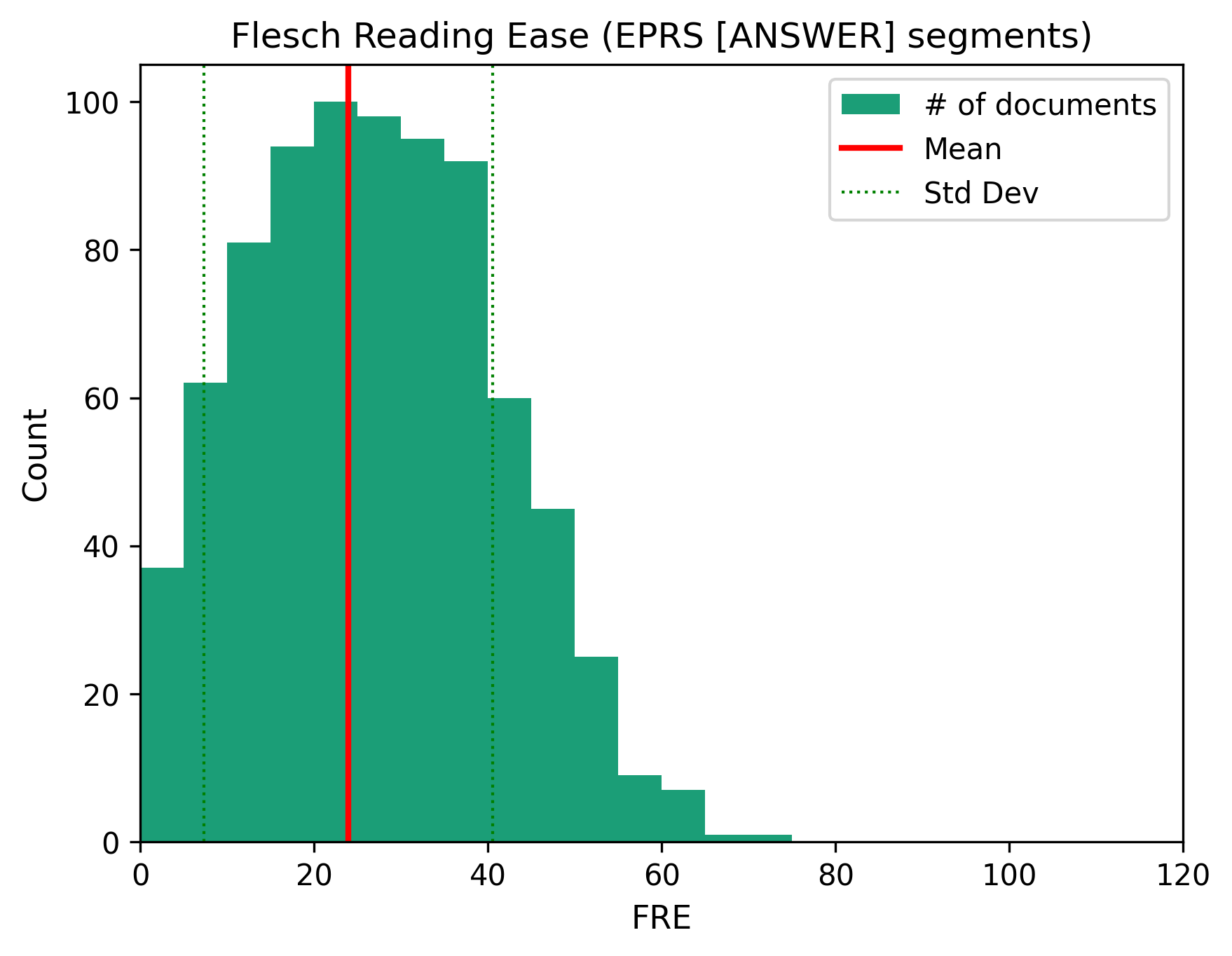}
    \caption{Readability (FRE)}
    \label{fig:conv_fre}
  \end{subfigure}

  \caption{Conversational dataset (\textit{EUDial}) statistics:
  (a) dialogue length in turns; 
  (b) EPRS answer segment lengths; 
  (c) question-to-answer ratio per adjacent citizen--EPRS pair; 
  (d) Flesch Reading Ease (FRE).}
  \label{fig:conv_grid}
\end{figure}

\paragraph{\textbf{EUDial Dialogue Dataset.}} Dialogues average 4.3 turns (min $=2$, max $=18$). Figure~\ref{fig:conv_turns_per_conv} shows the distribution, confirming that most dialogues fall between 6 and 12 turns. By role, citizen utterances average 12.0 words, and system answers average $39.7$ words. Figure~\ref{fig:conv_eprs_answer_lengths} highlights the distribution of system answers. Pairing each citizen turn with the subsequent EPRS yields a dialogue-level \textit{q$\to$a} ratio distribution with mean $=0.311$ (reflecting tighter, incremental exchanges compared to static QA pairs). Figure~\ref{fig:conv_q2a} illustrates the distribution across dialogues. Measured with Flesch Reading Ease (FRE), FRE averages $23.95$ per turn; when averaged per dialogue, the mean per-dialogue FRE is $23.89$, placing them in a difficult-to-read range for laypersons. Figure~\ref{fig:conv_fre} shows the FRE score distribution across answers.






\section{Prompt Templates}
In this section, we provide the prompt templates used for generating question and dialogue datasets. A one-shot prompt is designed for generating dialogue datasets.

\subsection{Question Generation Prompt}
\label{appendix:qg}
\begin{tcolorbox}[title=Question Generation Prompt, colback=gray!3, colframe=black!60, fonttitle=\bfseries,
breakable, listing only, listing options={basicstyle=\scriptsize\ttfamily, breaklines=true, columns=fullflexible}]
You are an AI assistant working for the Citizens’ Enquiries Unit (AskEP) within the European Parliamentary Research Service (EPRS).\\
\\
Your task is to infer the **citizen’s question** that the blog post is addressing, using the blog post title and the introductory paragraph provided.\\
\\
- Your response **must incorporate all important elements** mentioned in both the title and the paragraph.\\
- Avoid using abbreviations (e.g., write 'European Union' instead of 'EU').\\
- Do not add explanations, context, or commentary - respond with **only** the question.\\
- If no question is implied, reply only with: N/A\\
\\
`Title': ``\{title\}''\\
\\
`Paragraph': ``\{paragraph\}''\\
\\
What is the citizen’s question?
\end{tcolorbox}

\subsection{Dialogue Generation Prompt}
\label{appendix:dialoguegeneration}
\begin{tcolorbox}[title=Dialogue Generation Prompt, colback=gray!3, colframe=black!60, fonttitle=\bfseries,
breakable, listing only, listing options={basicstyle=\scriptsize\ttfamily, breaklines=true}]
\# Identity\\
\\
You are representing the European Parliamentary Research Service (EPRS).\\
Your role is to simulate an informative, multi-turn conversation with a citizen, grounded entirely in the answer provided in JSON format.\\
You must provide structured, informative, and neutral responses on citizen questions.\\
\\
\# Critical Constraints - Follow These Without Exception\\
\\
To ensure accuracy and compliance:\\
- You MUST base all content strictly on the provided answer JSON.\\
- You MUST NOT use external knowledge or hallucinate facts.\\
- You MUST NOT provide legal advice or personal recommendations.\\
- You MUST paraphrase all content by summarizing and rephrasing using different vocabulary and structure.\\
- You MUST NOT reuse phrases, clauses, or sentence patterns from the original text.\\
- Aim to convey the *gist* of each section in your own words, using simple and neutral explanations.\\
- You MUST NOT use markdown formatting (no triple backticks or code blocks).\\
- You MUST ONLY return a valid JSON array of objects (see Format section).\\
\\
\# Instructions\\
\\
Your goal is to guide the Citizen through the structured answer, one section at a time, using clear, paraphrased replies and purposeful follow-up questions that encourage deeper exploration.\\
\\
\#\# Conversation Design:\\
\\
1. **First turn**: Start with the **citizen's question**, verbatim.\\
2. **EPRS reply**:\\
- Provide a concise, paraphrased summary of the ``blog\_introduction'' section.\\
- End with an **informative follow-up question** inviting the citizen to explore another section.\\
3. **Subsequent turns**:\\
  - After each EPRS message, the citizen responds briefly.\\
  - EPRS then explains the next section, followed by another follow-up question.\\
4. **Final turn**:\\
  - Conclude by inviting further questions or offering a resource if relevant.\\
\\
\#\# Style \& Content Rules\\
\\
- Use **simple, neutral, and professional language**.\\
- Each EPRS reply must be **$\leq$ 60 words**.\\
- Include both tags in every EPRS utterance: [ANSWER] ... [FOLLOWUP QUESTION] ...\\
- Each EPRS turn must reference **exactly one section** (using the ``section'' field).\\
- Keep the dialogue coherent and aligned with EU informational tone.\\
\\
\# Format\\
\\
Return a JSON array. Each object must have:\\
- "role": either "citizen" or "eprs"\\
- "utterance": the message content\\
- "section": required only for "eprs" turns\\
\\
Do not include any other keys, commentary, or formatting.\\
\\
\# Example
\begin{verbatim}
[
  {{
    ``role'': ``citizen'',
    ``utterance'': ``What is the EU doing to help people
    access healthcare in another EU country?''
  }},
  {{
    ``role'': ``eprs'',
    ``utterance'': ``[ANSWER] That's an important question.
    The EU helps citizens access healthcare across borders by
    supporting national systems and ensuring legal rights to
    cross-border care. [FOLLOWUP QUESTION] Would you like to
    learn how scheduled treatments abroad are covered?'',
    ``section'': ``blog_introduction''
  }},
  {{
    ``role'': ``citizen'',
    ``utterance'': ``Yes, how are scheduled treatments
    covered?''
  }},
  {{
    ``role'': ``eprs'',
    ``utterance'': ``[ANSWER] If you're planning care in
    another EU country, you can be reimbursed or have your
    insurer pay directly, depending on the provider and
    authorisation process. [FOLLOWUP QUESTION] Shall I explain
    what happens if you need emergency care while abroad?'',
    ``section'': ``Planned healthcare''
  }}
]
\end{verbatim}

\# Input\\
\\
Citizen Question:\\
\{question\}\\
\\
Structured Answer (JSON):\\
\{answer\_json\}\\
\\
\# Output\\
\\
Only return the JSON array of conversation turns. Do not include markdown formatting, code blocks, or explanations. Follow the constraints exactly.
\end{tcolorbox}

\newpage
\section{Algorithms}
In this section, we provide the algorithms for generating multi-turn dialogue dataset from single-turn dialogues and proactive dialogue system.

\subsection{DialEU-AskEP Dialogue Dataset Algorithm}
\label{appendix:datasetalgorithm}

\begin{algorithm}[!ht]
\caption{Generate Multi-Turn Dialogue from a Single-Turn Dialogue}
\label{alg:dataset}
\begin{algorithmic}[1]
\Require Document collection $D=\{d_i\}_{i=1}^n$, where each $d_i=\langle q_i, a_i\rangle$,  
$q_i$ is the user question, $a_i=[s^{i}_{1},\ldots,s^{i}_{k}]$ is the multi-section answer,  
and each $s^{i}_{j}=\langle h^{i}_{j},c^{i}_{j}\rangle$ with $h^{i}_{j}$ the section header and $c^{i}_{j}$ the section content.  
\Ensure Dialogue dataset $T = \{T_1, \ldots, T_n\}$, where each $T_i$ is an ordered list of turns, and each turn is $\langle u^{i}_{j}, r^{i}_{j}, f^{i}_{j}\rangle$ consisting of the user utterance $u^{i}_{j}$,  
the system response $r^{i}_{j}$ grounded in $c^{i}_{j}$, and the follow-up $f^{i}_{j}$ pointing to $s^{i}_{j+1}$ (if any).
\For{each document $d_i \in D$}
    \State $T_i \gets \emptyset$ \Comment{initialize dialogue sequence}
    \State \textbf{// initial query turn}
    \State append $\langle \text{utterance}=q_i \rangle$ to $T_i$
    \For{$j \gets 1$ to $k$}
        \State \textbf{// response turn grounded in section } $s^{i}_{j}$
        \State $r^{i}_{j} \gets \textsc{Summarize}(c^{i}_{j})$ \Comment{concise paraphrase of $c^{i}_{j}$}
        \State $f^{i}_{j} \gets \textsc{FollowUp}(s^{i}_{j+1})$ \Comment{pointer to next section if any}
        \State append $\langle \text{utterance}=[\text{Answer}]\,r^{i}_{j}\,[\text{Follow-Up}]\,f^{i}_{j},\ \text{section}=h^{i}_{j} \rangle$ to $T_i$
        \If{$j<k$}
            \State \textbf{// acknowledgement turn}
            \State $u^{i}_{j+1} \gets \textsc{AcknowledgeOrQuery}(f^{i}_{j})$
            \State append $\langle \text{utterance}=u^{i}_{j+1} \rangle$ to $T_i$
        \EndIf
    \EndFor
    \State Add $T_i$ to $T$
\EndFor
\State \Return $T$
\end{algorithmic}
\end{algorithm}

\subsection{LexGuide Algorithm}
\label{appendix:dialeualg}

\paragraph{\textbf{Algorithm Details.}}
The algorithm has three phases:
\begin{enumerate}
    \item \textbf{Initialization and First Turn}: Retrieval via MMR, hierarchical topic tree construction, and response generation at the root node.
    \item \textbf{Dialogue Loop}: At each turn, if the user acknowledges the follow-up, traversal proceeds by BFS/DFS to children (default behavior). For new queries, the system checks (i) visited nodes $Y$, (ii) unexplored children $Z$, and (iii) global coverage in $G$. If none suffice, a new tree $G'$ is constructed.
    \item \textbf{Response and Navigation Updates}: The navigation state $S=(Y,L,Z,B,K,H)$ is updated, ensuring systematic exploration with flexibility for user-driven changes, backtracking, or topic shifts.
\end{enumerate}

\begin{algorithm}[!ht]
\caption{LexGuide: Hierarchical Topic-Guided Proactive Dialogue}
\label{alg:lexguide-framework}
\begin{algorithmic}[1]
\Require User query $q$, corpus $X$, fragments $P$, parameters $k$, $\lambda$, $\tau$
\Ensure Dialogue path $\mathcal{Y} = \{\langle u_1, r_1, f_1 \rangle, \ldots, \langle u_n, r_n, f_n \rangle\}$

\State \textbf{Initialization:}
\State Embed and index all fragments $p \in P$ into database
\State Initialize dialogue path $\mathcal{Y} \gets \emptyset$
\State Initialize navigation state $S = (Y=\emptyset, L=\text{null}, Z=\emptyset, B=\text{BFS/DFS}, K=[], H=[])$
\State Initialize topic tree $G \gets \text{null}$

\State \textbf{First Turn Processing:}
\State $R \gets \text{MMR-Retrieve}(q, P, k, \lambda)$ \Comment{Top-$k$ for initial answering}
\If{$|R| = 0$}
    \State \textbf{return} ``No relevant information found'', $\mathcal{Y}$
\EndIf
\State $G \gets \text{BuildHierarchy}(R)$ \Comment{Hierarchical Topic Modeling}
\If{$G = \text{null}$ \textbf{or} $|V| < 2$}
    \State Create single-node tree with all fragments in $R$
\EndIf
\State $L \gets \text{Root}(G)$ \Comment{Current node = root}
\State $Z \gets \text{Children}(L)$
\State $r \gets \text{GenerateResponse}(q, \text{Fragments}(L))$
\State $f \gets \text{GenerateFollowup}(Z, H)$
\State $\mathcal{Y} \gets \mathcal{Y} \cup \{\langle q, r, f \rangle\}$

\State \textbf{Dialogue Loop:}
\While{user continues}
    \State $u \gets \text{GetUserInput}()$ \Comment{With timeout handling}
    \If{timeout occurred}
        \State \textbf{break} \Comment{Inactivity termination}
    \EndIf
    
    \If{$\text{UserAcknowledges}(u, f)$} \Comment{Follow system suggestion}
        \State $q \gets f$
        \State $L_{\text{next}} \gets \text{SelectCluster}(f, Z, B)$
        \State $\text{fragments} \gets \text{Fragments}(L_{\text{next}})$
        
    \Else \Comment{User provides new query $q'$}
        \State $q \gets u$
        
        \Comment{Step 1: Try resolving within visited nodes $Y$}
        \State $L_{\text{next}} \gets \arg\max_{g \in Y} \text{Sim}(q, C_g)$
        \If{$\text{Sim}(q,C_{L_{\text{next}}}) \geq \tau$}
            \State $\text{fragments} \gets \text{Fragments}(L_{\text{next}})$
        \Else
            \Comment{Step 2: Try unresolved children $Z$ of current node $L$}
            \State $L_{\text{next}} \gets \arg\max_{g \in Z} \text{Sim}(q, C_g)$
    \algstore{lexguide}
\end{algorithmic}
\end{algorithm}

\clearpage 

\begin{algorithm}[!ht]
\caption{LexGuide: Hierarchical Topic-Guided Proactive Dialogue (Part 2)}
\label{alg:lexguide-framework-part2}
\begin{algorithmic}[1]
\algrestore{lexguide}
            \If{$\text{Sim}(q,C_{L_{\text{next}}}) \geq \tau$}
                \State $\text{fragments} \gets \text{Fragments}(L_{\text{next}})$
            \Else
                \Comment{Step 3: Global coverage over entire tree $G$}
                \State $\text{coverage} \gets \max_{g \in G} \text{Sim}(q, C_g)$
                \If{$\text{coverage} \geq \tau$}
                    \State $L_{\text{next}} \gets \arg\max_{g \in G} \text{Sim}(q, C_g)$
                    \State $\text{fragments} \gets \text{Fragments}(L_{\text{next}})$
                \Else
                    \Comment{Outside scope: build new tree $G'$}
                    \State $R \gets \text{MMR-Retrieve}(q, P, k, \lambda)$
                    \State $G \gets \text{BuildHierarchy}(R)$
                    \State $L_{\text{next}} \gets \text{Root}(G)$
                    \State $\text{fragments} \gets \text{Fragments}(L_{\text{next}})$
                    \State Reset navigation state $S$
                \EndIf
            \EndIf
        \EndIf
    \EndIf

    \State \textbf{Response Generation and Navigation Update:}
    \State $r \gets \text{GenerateResponse}(q, \text{fragments})$
    \State $Y \gets Y \cup \{L\}$ \Comment{Update visited nodes}
    \State $L \gets L_{\text{next}}$ \Comment{Advance current node}
    \State $K \gets K + [L]$ \Comment{Update path}
    \State $Z \gets \text{UpdateChildren}(G, L, Y)$ \Comment{Unexplored children}
    \State $H \gets H + [(u,r)]$ \Comment{Update history}
    
    \If{$Z \neq \emptyset$}
        \State $f \gets \text{GenerateFollowup}(Z, H)$
    \Else
        \State $f \gets \text{null}$
    \EndIf
    
    \State $\mathcal{Y} \gets \mathcal{Y} \cup \{\langle u, r, f \rangle\}$
\EndWhile

\State \textbf{Termination:}
\If{all relevant nodes visited}
    \State \textbf{return} ``All relevant topics explored'', $\mathcal{Y}$
\ElsIf{user inactivity or explicit termination}
    \State \textbf{return} ``Dialogue terminated by user'', $\mathcal{Y}$
\EndIf
\end{algorithmic}
\end{algorithm}

\end{document}